\documentclass[letterpaper, 10 pt, conference]{ieeeconf}
\IEEEoverridecommandlockouts
\usepackage[utf8]{inputenc}
\usepackage{cite}
\usepackage{amsmath,amssymb,amsfonts}
\usepackage{algorithmic}
\usepackage{graphicx}
\usepackage{textcomp}
\usepackage{url}
\usepackage{subcaption}
\usepackage[acronym]{glossaries}
\usepackage[normalem]{ulem}
\usepackage{amsfonts}
\usepackage{booktabs}
\usepackage{multirow}
\usepackage[table]{xcolor}
\usepackage{array}
\usepackage{bytefield}
\usepackage{float}
\usepackage{gensymb}

\newacronym{dvs}{DVS}{Dynamic Vision Sensor}
\newacronym{cnn}{CNN}{Convolutional Neural Network}
\newacronym{gpu}{GPU}{Graphics Processing Unit}
\newacronym{lidar}{LIDAR}{Light Detection and Ranging}
\newacronym{radar}{RADAR}{Radio Detection And Ranging}
\newacronym{tof}{ToF}{Time of Flight}
\newacronym{sgm}{SGM}{Semi-Global Matching}
\newacronym{cis}{CIS}{Conventional CMOS-based sensors}
\newacronym{led}{LED}{Light-emitting diode}

\newcommand{\targetspeed}{$15\,\text{m/s}$}
\newcommand{\targetdiameter}{$2\,\text{cm}$}
\newcommand{\nerfmaxspeed}{{$23.4\,\text{m/s}$}}
\newcommand{\nerfminspeed}{{$16.0\,\text{m/s}$}}
\newcommand{\viconmaxspeed}{{$4.8\,\text{m/s}$}}
\newcommand{\viconminspeed}{$1.5\,\text{m/s}$}

\newcommand{\dartdegerror}{$24.73\degree$}
\newcommand{\dartraderror}{$18.4\,\text{mm}$}
\newcommand{\collisionerror}{$25.03\%$}
\newcommand{\predictionrate}{$3\,\text{ms}$}
\newcommand{\predictionratetwo}{$1\,\text{ms}$}

\begin{document}

\title{\LARGE \bf
Fast Motion Understanding with Spatiotemporal Neural Networks\\
and Dynamic Vision Sensors
}

\author{Anthony Bisulco, Fernando Cladera Ojeda, Volkan Isler, Daniel D. Lee% <-this % stops a space
\thanks{All authors are with the Samsung AI Center NY, 837 Washington Street, New York, New York 10014}%
}

\maketitle

\thispagestyle{empty}
\pagestyle{empty}

%%%%%%%%%%%%%%%%%%%%%%%%%%%%%%%%%%%%%%%%%%%%%%%%%%%%%%%%%%%%%%%%%%%%%%%%%%%%%%%%
\begin{abstract}
This paper presents a \gls{dvs} based system for reasoning about high speed motion. As a representative scenario, we consider the case of a robot at rest reacting to a small, fast approaching object at speeds higher than \targetspeed{}.
Since conventional image sensors at typical frame rates observe such an object for only a few frames, estimating the underlying motion presents a considerable challenge for standard computer vision systems and algorithms.
In this paper we present a method motivated by how animals such as insects solve this problem with their relatively simple vision systems.

Our solution takes the event stream from a \gls{dvs} and first encodes the temporal events with a set of causal exponential filters across multiple time scales. We couple these filters with a \gls{cnn} to efficiently extract relevant spatiotemporal features.
The combined network learns to output both the expected time to collision of the object, as well as the predicted collision point on a discretized polar grid.
These critical estimates are computed with minimal delay by the network in order to react appropriately to the incoming object.
We highlight the results of our system to a toy dart moving at \nerfmaxspeed{} with a \dartdegerror{} error in $\theta$, \dartraderror{} average discretized radius prediction error and \collisionerror{} median time to collision prediction error.

\end{abstract}

%%%%%%%%%%%%%%%%%%%%%%%%%%%%%%%%%%%%%%%%%%%%%%%%%%%%%%%%%%%%%%%%%%%%%%%%%%%%%%%%
\section{INTRODUCTION}

Practical mobile robotic systems need to operate in a wide range of environmental conditions including farms~\cite{Pretto_2020}, forests~\cite{chen2019sloam}, and caves~\cite{tabib2020autonomous}.
These dynamic environments contain numerous fast moving objects such as falling debris and roving animals that mobile robots need to sense and respond to accordingly.
The performance criteria for operating in such scenarios require sensors that have fast sampling rates, good spatial resolution and low power requirements.
Active sensor solutions such as structured light or \gls{tof} sensors fail to meet this criteria due to their high power consumption and limited temporal resolution~\cite{wang2017characterization, Zaffar_2018}.
An alternative approach is to track objects with conventional cameras by explicitly computing dense optical flow in time. However, these algorithms require computing dense feature correspondences and the heavy processing requirements limit overall system performance~\cite{OpticalFlowHistoric}.

\glsreset{dvs} \glspl{dvs}
are biologically-inspired vision sensors that provide near continuous sampling and good spatial resolution at low power.
Motivated by the ability of insects to efficiently respond to fast motion stimuli with their simple visual systems, this work investigates the use of efficient and economical neural networks to process the event streams from \glspl{dvs} to respond to fast moving objects. Insects such as Drosophilia are able to quickly and accurately react to fast approaching objects via recognition of critical spatiotemporal motion patterns~\cite{drosophilia}; thus, we seek to construct an efficient vision system that is able to respond to moving objects with speeds greater than \targetspeed{}, and with diameters as small as \targetdiameter{}.

A \gls{dvs} sensor measures asynchronous changes in light intensity~\cite{Gallego_2020} rather than performing full synchronous readouts of the pixel array as with
conventional CMOS-based sensors.
A \gls{dvs} outputs a stream of events $E=\{e_i\}$ where
each event can be described as a vector $e_i= (x, y, p, t)$. $x$ and $y$ describe the image pixel location, $p$ is for polarity which is the sign of the light change, and $t$ is the time at which the event occurs.

\begin{figure}[t]
  \centering
  \includegraphics[width=.9\linewidth]{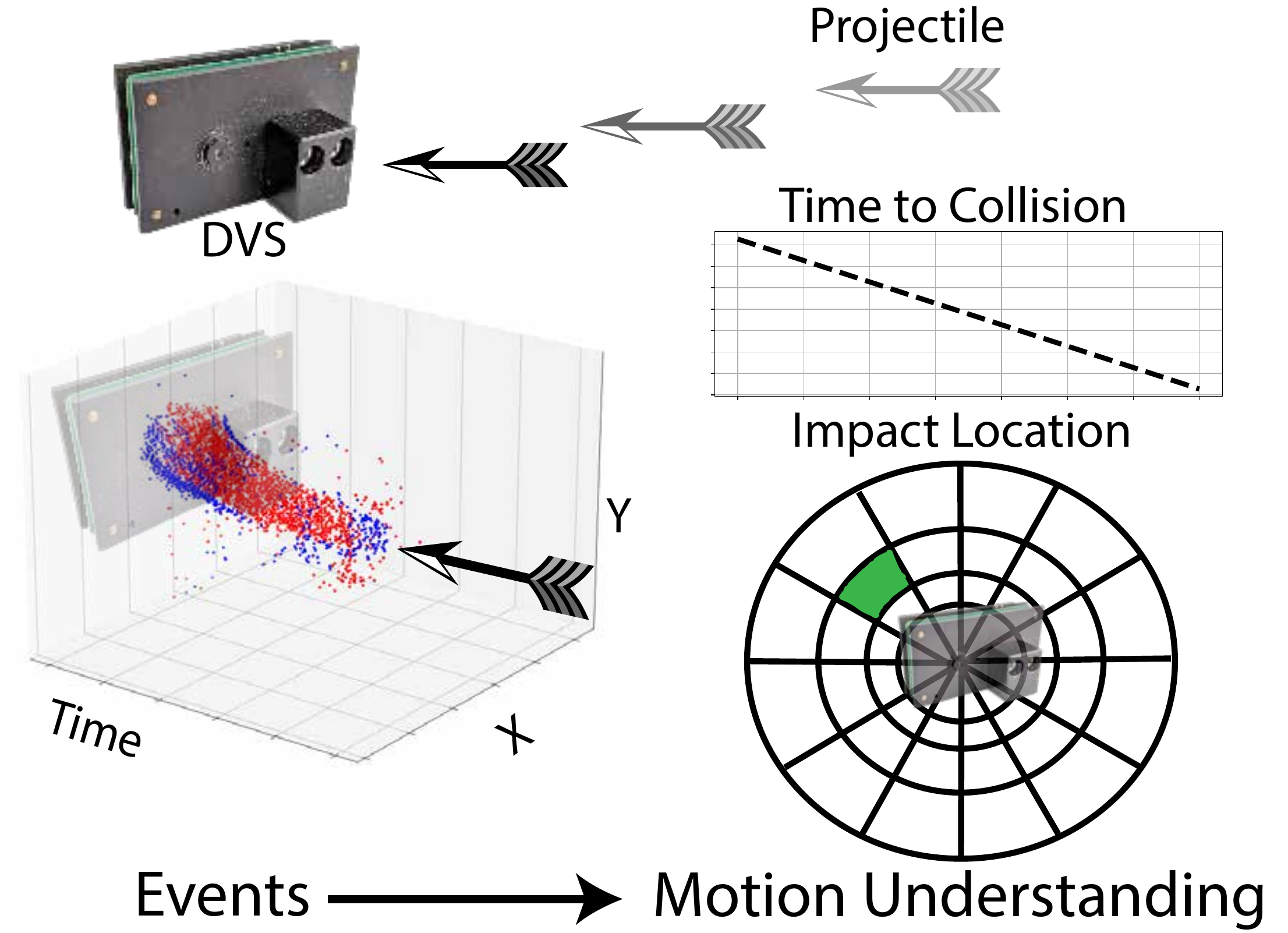}
  \caption{Motion understanding of time to collision and collision location of fast moving objects using the \gls{dvs} spatiotemporal event volume. Our system outputs the coordinates of the impact on a polar grid centered at the camera, as well as the time to collision to the camera plane.}
  \label{fig:f1}
\end{figure}

We address the problem of predicting the time to collision and impact location of a fast approaching object from the \gls{dvs} event stream.
Our solution consists of encoding motion and extracting relevant spatiotemporal features using a bank of exponential filters with a \gls{cnn}.
We show that our method is able to yield robust estimates for the predicted time and location of the future impact without needing to compute explicit feature correspondences (Fig.~\ref{fig:f1}).
We validate the performance of our system on fast moving objects such as a dropped ball and shooting dart (from a toy gun) and quantify its overall performance in terms of accuracy and latency.

The overall contributions of this work include the following: 1) a novel network architecture combining a temporal filterbank with a \gls{cnn} to efficiently learn spatiotemporal patterns in the \gls{dvs} event stream, and 2) datasets and a new training protocol that uses motion capture based 3D motion information, and augments the measured \gls{dvs} events by warping the object trajectories to generate new event streams for training the network.

%%%%%%%%%%%%%%%%%%%%%%%%%%%%%%%%%%%%%%%%%%%%%%%%%%%%%%%%%%%%%%%%%%%%%%%%%%%%%%%%

\section{Related Work}
\label{sec:relwork}
Prior work on collision detection for autonomous mobile robots have utilized a number of exteroceptive sensors.
Active sensors can achieve unmatched levels of performance, enabling long distance detection of obstacles. For example,
\gls{tof} sensors, such as \gls{lidar}, precisely measure the time difference between the emission of a directed pulse and its reflection from the target surface and can be used to accurately track a moving object.
Unfortunately, their weight and power requirements preclude their comprehensive use on small autonomous systems~\cite{obstacleAvoidanceRadar}\cite{ousterLidar}.

Passive vision-based sensors are attractive for mobile robots but accurate tracking of moving objects requires correlating multiple measurements over time.
One conventional approach uses multiple cameras to first compute disparity maps and then track 3-D features over time to estimate object motion. Computing dense correspondences over time and space requires specialized computing resources~\cite{SGMFpga} or mitigating the computational load with lower resolution sensors~\cite{stereoVisionQuad}.
With monocular cameras, assumptions about scale~\cite{monocularVision}, or obtaining additional information from auxiliary sensors is required.  Integrating this information using conventional computer vision techniques typically requires additional computational resources for real-time motion estimation~\cite{davideMicroDrone}.

Recent work has demonstrated the utility of \gls{dvs} sensors for collision detection.
In~\cite{ballDetection}, the problem of ball detection and tracking was analyzed. The authors proposed a Hough transform-based approach to detect and track balls, reaching frequencies of up to $500\,\text{Hz}$.
This approach was specifically designed for the case of a ball and has not been shown to generalize to other settings.
More recently, EVDodgeNet~\cite{evdodgeNet} proposed a full learning-based approach to detect quadrotor orientation, segment events, and estimate obstacle approach. As an effective end-to-end solution, this approach has two limitations: a perceptual lag of approximately $60\,\text{ms}$ and its complexity and computational overhead, requiring three separate networks to estimate the homography, deblur the image, and segment the tracked object. Additionally, this work relies on accurate simulators to generate sufficient training data for training the network and may suffer from unmodeled differences in the physical system~\cite{zhu2019eventgan}. Recently, Falanga \emph{et al.} showcased an obstacle avoidance method for quadrotors in~\cite{davideObstacleAv}, relying on ego-motion compensation and tracking to estimate the velocity of the incoming object.
In contrast, our proposed system is much simpler in architecture yet is able to generate accurate motion estimates that are sufficient for robotic systems to respond in real-time to faster moving objects.

\section{Method}
\begin{figure}
  \centering
  \includegraphics[width=\linewidth]{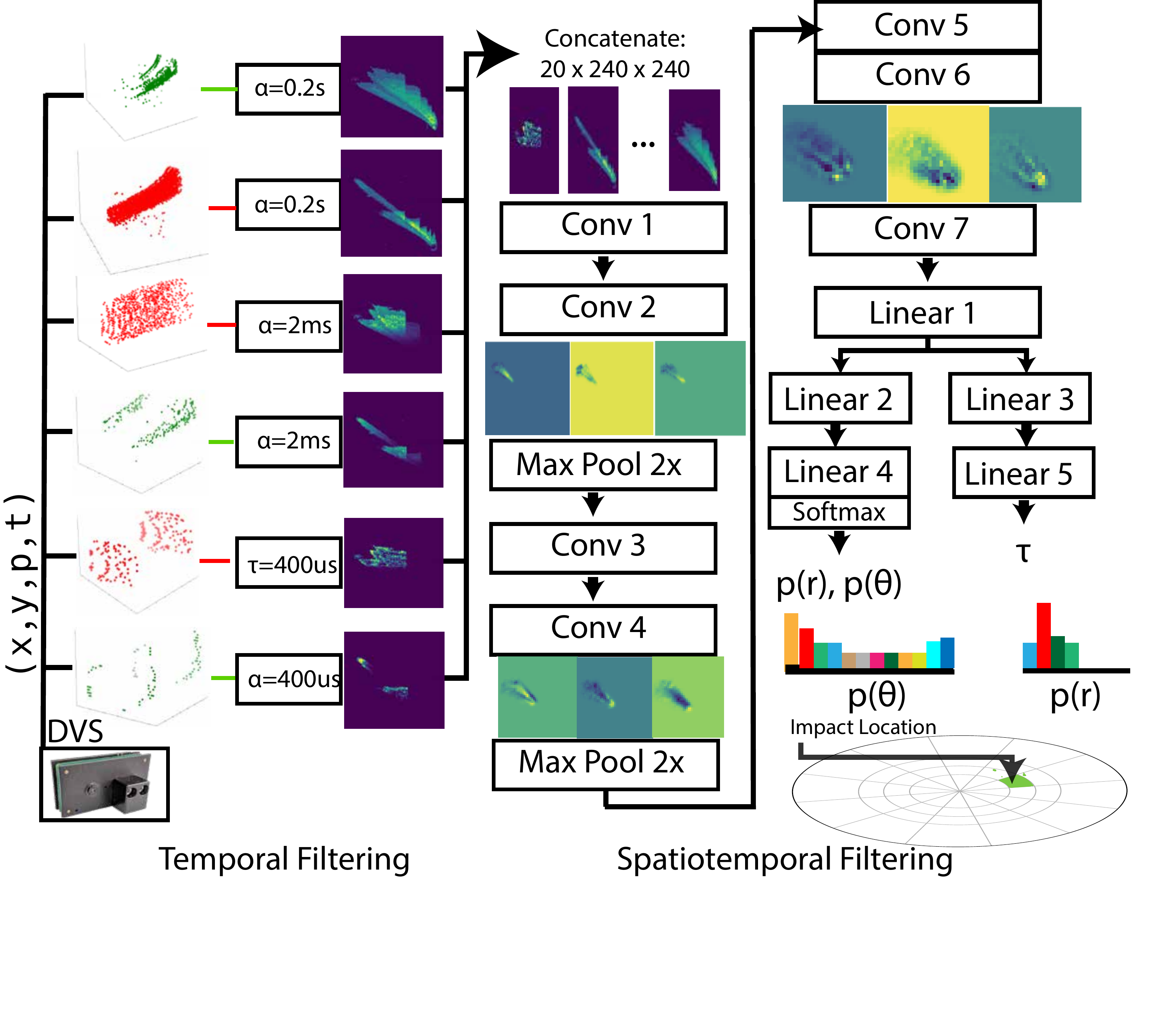}
  \caption{End to end architecture: 1)~Temporal filter containing a bank of exponential filters integrating over different time lengths. 2)~Spatiotemporal filters performed using the \gls{cnn} architecture which outputs time to collision and the coordinates of the impact on a polar grid.}
  \label{fig:app}
\end{figure}

We present our method for predicting time to collision~$\tau$ and object impact location in polar coordinates~$(R, \theta )$  using a \gls{dvs} camera for fast moving objects.
Our method consists of various stages including:
1) a bank of exponential filters for encoding object motion, and
2) a \gls{cnn} for extracting spatiotemporal features,  predicting time to collision and predicting a discrete polar output location.
Supervised training is performed on the \gls{cnn}, thus we collected a  dataset of object state and \gls{dvs} events for training. To expand our dataset and enhance the performance of our \gls{cnn} we used a set of novel augmentations applied to the \gls{dvs} events.

\subsection{Data Collection}

\begin{figure}
  \centering
  \includegraphics[width=\linewidth]{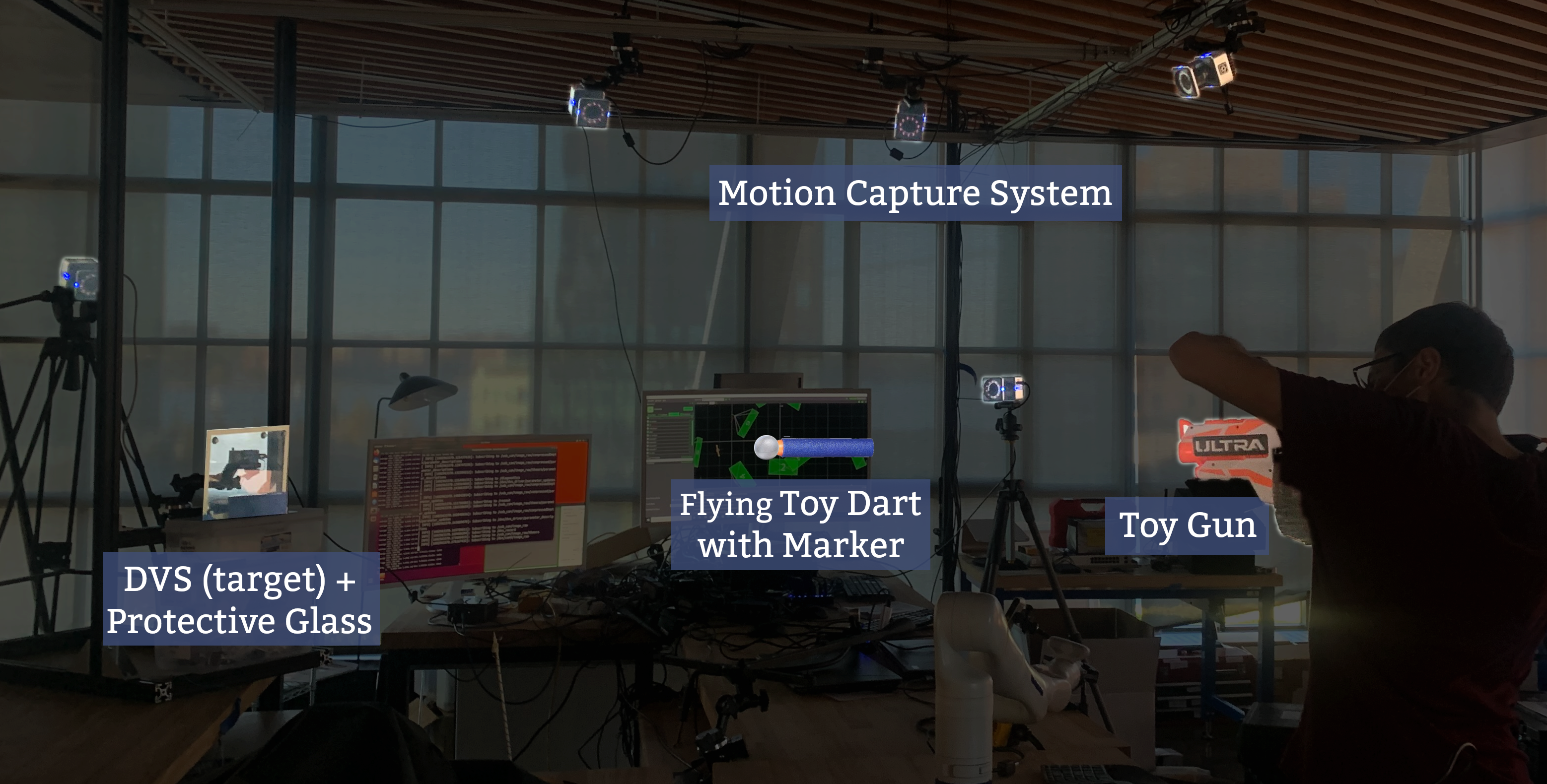}
  \caption{Experimental setup for data acquisition for the toy dart dataset.
  Darts are tracked using a motion capture system while they are recorded with
  the \gls{dvs}. To ensure proper temporal synchronization white and IR
  \glspl{led} are blinked before each shot. Spatial calibration using the array
  of \glspl{led} is performed at the beginning of the data collection run.}
  \label{fig:datacollection}
\end{figure}

The data collection process consists of acquiring a series of \gls{dvs} recordings while objects are  approaching the sensor. In this study, we targeted two objects: falling balls and toy darts.
The falling ball dataset was obtained by dropping spherical markers to a
\gls{dvs}. The toy dart dataset was obtained by shooting darts to a \gls{dvs}
(Fig.~\ref{fig:datacollection}).

In both cases, we are interested in finding the spatial coordinates where the object will impact with the plane where the camera is located, as well as the required time to collision.
Our coordinate system defines +Z to be perpendicular to the plane of the image sensor, and we define the origin in the camera itself.
Therefore, collision location and collision time can be extracted when $Z_{object} = 0$ (when the object is located in the plane of the camera). After calculating the impact location, we use collision time  to calculate time to collision for each previously indexed location $i$ as $\tau=T_{i}-T_{impact}$.
We tracked the motion of the objects using a VICON motion capture system with reflective markers.

To ensure proper tracking and synchronization between the data recorded with the \gls{dvs} and the motion capture system, we performed both spatial and temporal calibration of our system. Spatial calibration was performed using an $7\times5$ array of \glspl{led} blinking at $2\,\text{Hz}$ similar to the one described in~\cite{cameraCalib}. We attached markers to the calibration board, to track its position while we performed the calibration of the camera. By doing this, we can find the transformation between the camera and the motion capture system ${}^{MoCap}_{}\mathcal{T}_{DVS}$, as well as the camera intrinsics.

Temporal calibration was performed by blinking synchronized infrared and white \glspl{led} that were captured by both the motion capture system as well as the \gls{dvs}. This allows us to calculate the offset between these clocks at the beginning of each experiment.
The drift between both clocks after synchronization is negligible because each of our experiments only run for a few seconds.

This data collection procedure was performed to collect 150 ball drops and 36 toy dart shots. Our network is trained for each individual object for which we split the data into $80\%$ training set and $20\%$ testing set.  Speed ranges for the ball were {\viconminspeed{}-\viconmaxspeed{}} and toy dart were {\nerfminspeed{}-\nerfmaxspeed{}}. Initial drop height of the ball ranged from  $0.4$ to $1.2\,\text{m}$ and of the toy dart $0.6$ to $1\,\text{m}$. Trajectory time length for the ball ranged from $183\,\text{ms}$ to $352\,\text{ms}$ and for the toy dart from $26\,\text{ms}$ to $46\,\text{ms}$.

\subsection{Processing}

A key aspect of our architecture is a bank of exponential filters which integrate a signal in a time windowed region with an exponential kernel:
\begin{align}
y[n]=\alpha y[n-1]+(1-\alpha)x[n]
\end{align}
where $y[n]$ is the output signal, $x[n]$ is the input signal,  $\alpha$ is the smoothing factor  and $0<\alpha<1$ and n is a unit of discrete time.
We designed 10 different exponential filters$(\alpha_i)$ that encode motion over periods of $200\,\mu\text{s}$, $477\,\mu\text{s}$, $1.13\,\text{ms}$,
$2.71\,\text{ms}$, $6.47\,\text{ms}$, $15.44\,\text{ms}$,
$36.84\,\text{ms}$, $87.871\,\text{ms}$, $0.2\,\text{s}$, and $0.5\,\text{s}$ (Fig.~\ref{fig:app}).

Our exponential filter is implemented for each pixel on the image grid.
Separate filters are applied to the positive and negative polarity channels creating 20 $240 \times 240$ outputs  per time step.
Each filter is updated every $200\,\mu\text{s}$, which is our time step.  Outputs of the exponential filter every $200\,\mu\text{s}$ are temporally correlated, hence we regulate the filter to output every $3\,\text{ms}$ for the ball and every $1\,\text{ms}$ for the toy dart.
After the temporal downsampling, we perform a $2 \times$ spatial downscale and center crop.
Additionally, we linearly discretize the filter output for a reduced memory footprint to an unsigned integer ($8\,\text{bits}$).

\subsection{Augmentation}

\begin{figure}[b]
  \centering
  \includegraphics[width=.8\linewidth]{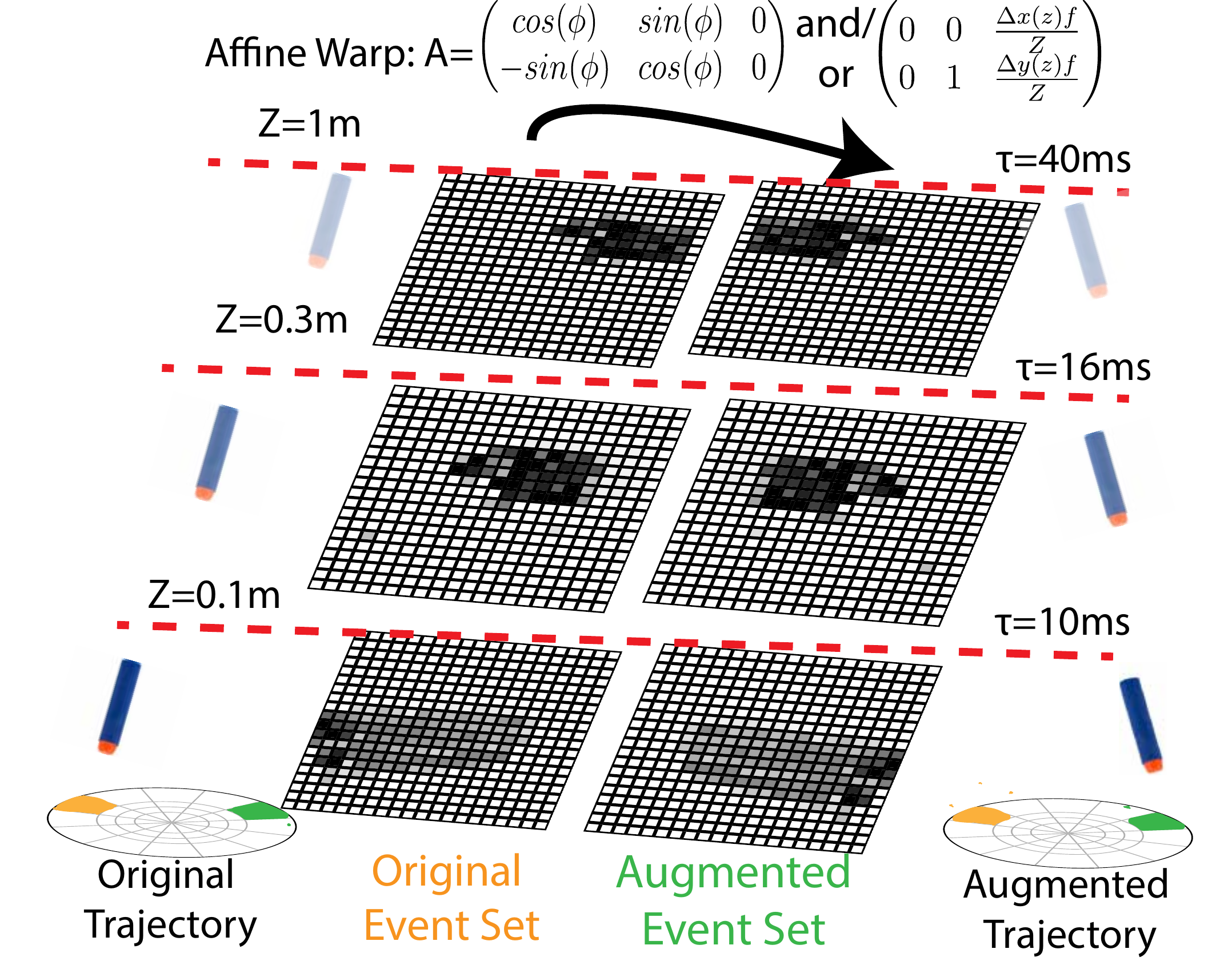}
  \caption{Augmentation procedure for shifting impact location or rotation about the optical axis. Translation augmentations perform a translation per depth value on the windowed event stream. Rotation augmentations perform an affine rotation on the windowed event stream. }
  \label{fig:augs}
\end{figure}

We perform a series of augmentations to ensure that the training set is balanced across the network output. However, care must be taken when moving the impact location since the amount of shift varies across the sequence as it depends on the depth due to motion parallax. Since we have ground truth from motion capture system, we are able to generate a series of novel event level augmentations which we describe next (Fig.~\ref{fig:augs}).

\textbf{Shifting the impact location:} We pick a random perturbation of the impact location given as a translation.  For each event time window, we translate the object events in the corresponding depth plane and compute the resulting event translation by projecting back on to the image plane.

\textbf{Rotation around the optical axis:} This transformation can be expressed as a homography independent of depth which allows us to transform each set of events accordingly. We generate new data by picking a rotation uniformly at random.

\subsection{Network}
The tensor of dimensions  $20\times240\times240$ issued from the exponential filters is used as the input of our network.
The next step in our processing pipeline is to perform spatiotemporal feature extraction. We emulate spatiotemporal feature extraction through convolution kernels to interpret the motion volume. Our network architecture consists of 7 convolution layers, two $2 \times$ max pooling layers and two readout networks, one for time to collision and one for $\left(p(r),p(\theta)\right)$ estimation.  $\theta$ is discretized into 12 bins each 30\degree and $R$ is discretized into 4 bins corresponding to:
\{$0\,\text{mm}$-$60\,\text{mm}$,
$60\,\text{mm}$-$91\,\text{mm}$,
$91\,\text{mm}$-$121\,\text{mm}$,
$121\,\text{mm}$-$\infty$\}

Our network is trained using the Adam optimizer with a learning rate of $10^{-4}$ and a batch size of 160. The loss function consists of the multi-objective loss  of time to collision mean squared error, $p(\theta_g)$ cross entropy loss and $p(r_l)$ cross entropy loss.
We use the exponential linear unit as our activation function between layers.
Our network is trained for a specific object and evaluated using the network for this object motion.

%%%%%%%%%%%%%%%%%%%%%%%%%%%%%%%%%%%%%%%%%%%%%%%%%%%%%%%%%%%%%%%%%%%%%%%%%%%%%%%%

\section{Results}

%%%%%%%%%%%%%%%%%%%%%%%%%%%%%%%%%%%%%%%%%%%%%%%%%%%%%%%%%%%%%%%%%%%%%%%%%%%%%%%%

\begin{figure*}
     \centering
     \begin{subfigure}[b]{0.49\textwidth}
        \centering
        \includegraphics[width=\linewidth]{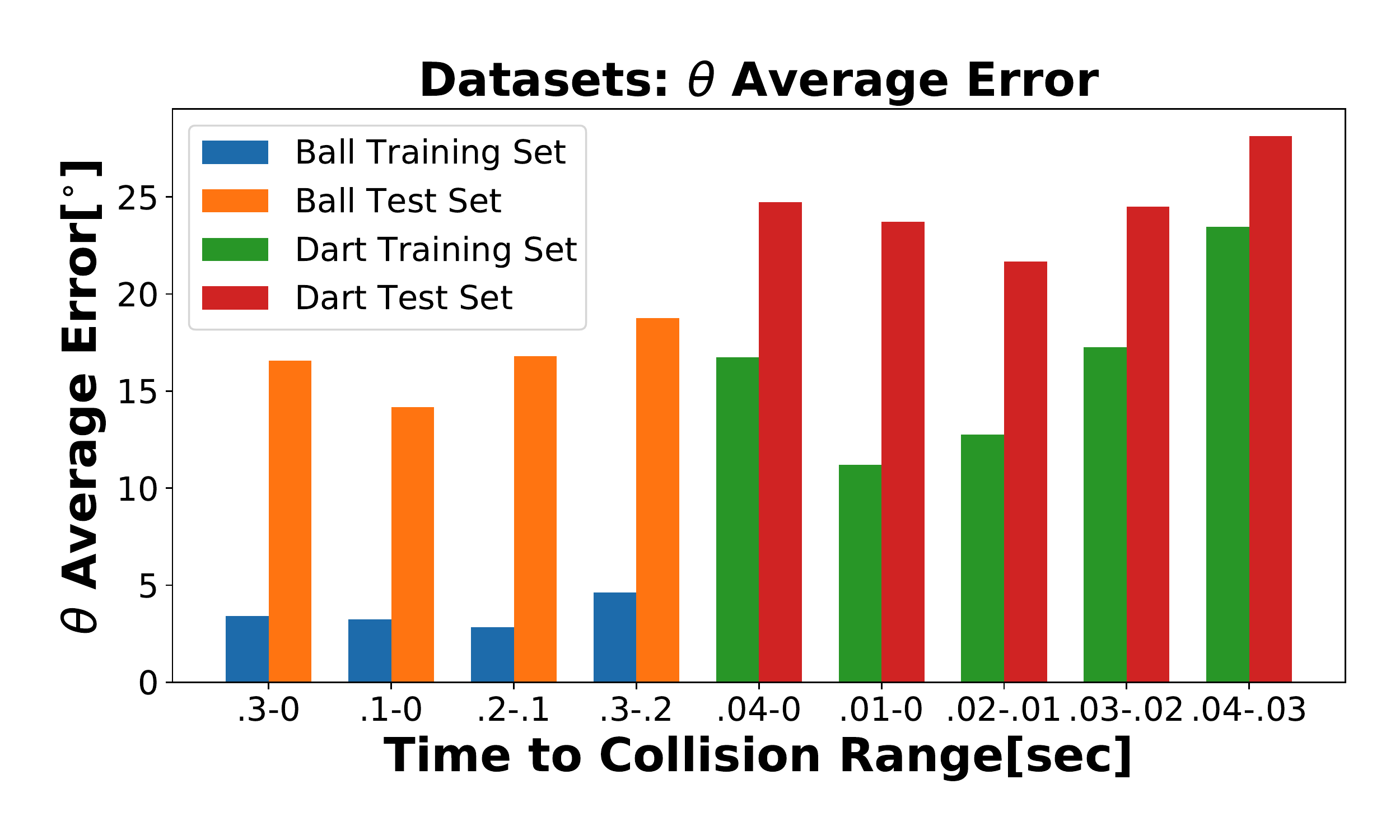}
        \caption{$\theta$ prediction average error.}
        \label{fig:com}
     \end{subfigure}
     \hfill
     \begin{subfigure}[b]{0.49\textwidth}
        \centering
        \includegraphics[width=\linewidth]{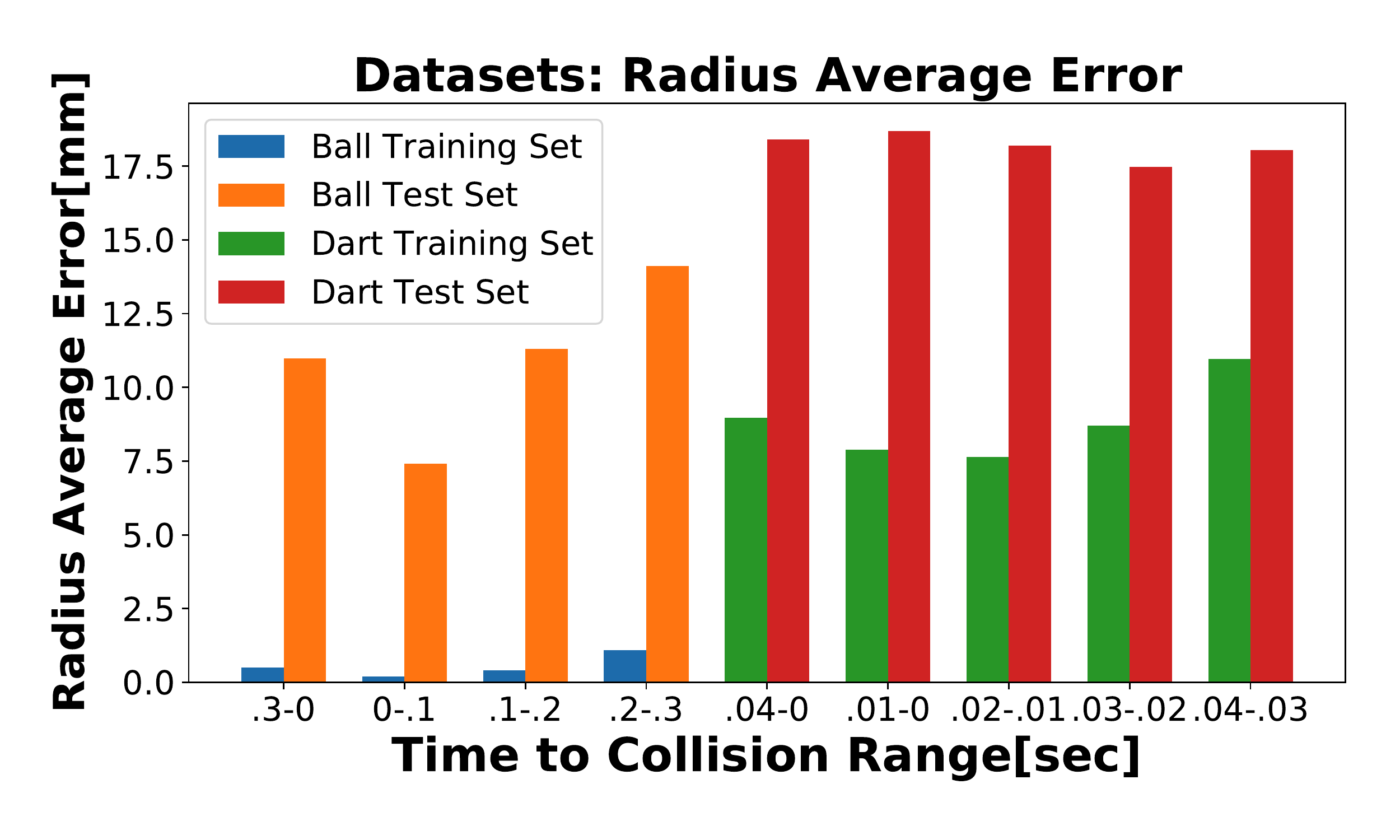}
        \caption{$R$ prediction average error.}
        \label{fig:comr}
     \end{subfigure}
        \caption{Training and test results for various time to collision bins for the dart and ball datasets. Both datasets show a trend of increased performance towards lower time to collision values. This suggests a speed/accuracy tradeoff present in our system.
        }
        \label{fig:res}
\end{figure*}

\begin{figure}[t]
  \centering
  \includegraphics[width=\linewidth]{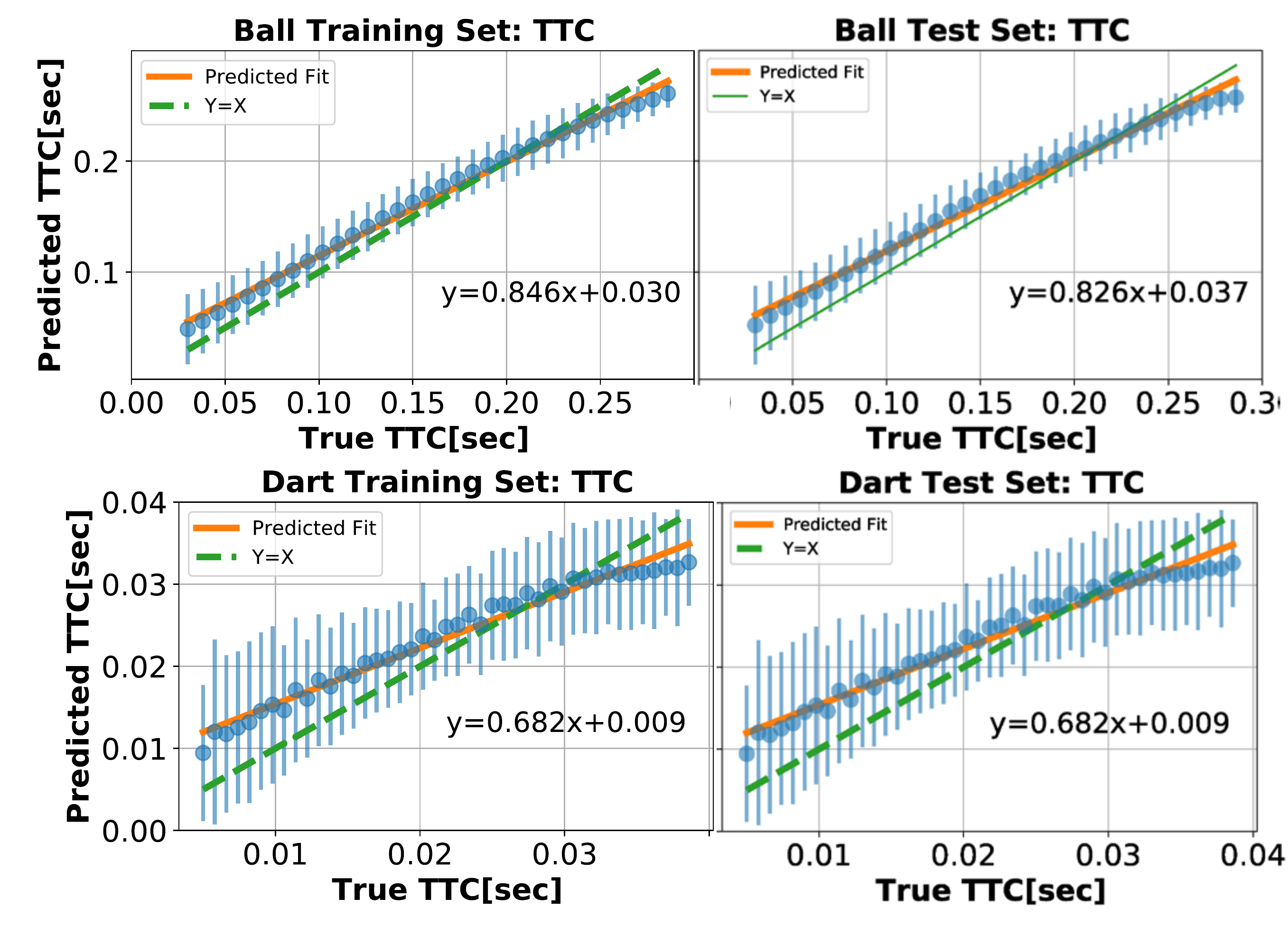}
  \caption{Time to collision correlation between predictions and ground-truth. We observe a very accurate correlation for the ball dataset, whereas the higher speeds in the dart dataset produce noisier estimations.}
  \label{fig:resultsttc}
\end{figure}

We highlight the results on the performance of our network on two datasets:
the ball with a max speed of \viconmaxspeed{},  and the dart dataset with a max speed of \nerfmaxspeed{}.
Our evaluation criteria for time to collision consists of the percent error between the true and estimated time to collision
\begin{align}
    E_{ttc, k}=\frac{|\tau_{k}- \hat \tau_{k}|}{\tau_{k}}
\end{align}
for all $k$ data samples. We take the median of this estimate across all time steps for robustness to outlier values from small time to collision predictions.  Our criteria for discrete theta prediction is the loss
\begin{align}
E_{\theta, k}=30 |\theta_{k}- \hat \theta_{k}|^2_2\quad  \{ \theta_k  \in \mathbb{Z}_{0}^{+} | \theta_k<12 \}
\end{align}
where 30 corresponds to the discrete loss per bin,
and radius is the loss
\begin{align}
    E_{r, k}=|f(r_k)-f(\hat r_k)|^2_2 \quad \{r \in \mathbb{Z}_{0}^{+} | r<4 \}
\end{align}
where f is the minimum value of the range for bin $r_k$.
Additionally, we highlight  various network components and their impact on algorithm performance. We predict on the exponential filter output every \predictionrate{} for the ball and every \predictionratetwo{} for the dart both the time to collision $\tau$ and discrete polar  impact location $(R, \theta)$.

\subsection{Ball Dataset}

The ball dataset consisted of a training set of 18,210 augmented examples and a testing set of 4,228 augmented examples.
Each example consists of the ball in a new starting position and impact point. Our performance on the training set for $(R, \theta)$ is an average $\theta$ error of $3.41\degree$ and an average radius error of $0.5\,\text{mm}$ (Fig.~\ref{fig:res}).
This highlights our algorithm is accurately able to predict $(R, \theta)$ throughout the trajectory length. Inspecting the testing results we see the average $\theta$ error of $16.56\degree$ and an average radius error of $10.99\,\text{mm}$.
These errors are larger than the training set, although still within the discretization interval for each quantity. Hence, we are still able to generalize to new trajectories of the same object for predicting $(R, \theta)$ over the trajectory length.
The median percent time to collision error globally for the training and testing set respectively were $9.58\%$ and $10.23\%$. We inspect our range of collision prediction in Fig.~\ref{fig:resultsttc} and showcase our close match to the true time to collision value.

We analyze the prediction results for  $(R, \theta)$  during different time intervals to evaluate the performance of our system when receiving an incoming stream from the sensor.
For this, we evaluated our algorithm
throughout the time to collision intervals of  $0.3\,\text{s}$-$0.2s$, $0.2\,\text{s}$-$0.1\,\text{s}$ and $0.1\,\text{s}$-$0\,\text{s}$. The results for both the training and testing set are within a discretization interval, demonstrating good performance.
One interesting pattern in the testing set ($R, \theta$) error is we see performance enhances as the object gets to smaller time to collision values.
This is probably due to objects strong spatiotemporal features that are present in small time to collision values rather than large time to collision values.
This result also highlights a speed-accuracy trade off present in our model.
One can get less accurate estimates at higher time to collision values or wait for longer and get more accurate results.
This is an inherent trade off in our system to quickly react to incoming obstacles.

\subsection{Toy Dart Dataset}

The dart dataset consisted of a training set of 6000 augmented training examples and 2910 augmented testing examples. Our performance on the training set for $(R, \theta)$ errors are $8.97\,\text{mm}$ and $16.75\degree$.
This value is larger than the error on the ball. Nonetheless, the dart is moving at much faster speed, showcasing the previously discussed speed-accuracy trade-off.
The results on the testing set for $(R, \theta)$ errors are \dartraderror{} and $ \dartdegerror{}$ (Fig.~\ref{fig:res}).
Again, both the training set and testing set are within the discretization interval demonstrating good algorithm performance over the trajectory length even for a fast moving object.
The median percent time to collision error globally for the training and testing set respectively were $16.75\%$ and $24.73\%$.
We inspect our model over the range of time to collision values in Fig. \ref{fig:resultsttc} and note the bias at the end and beginning of the trajectory.
This bias is due to limited event in the beginning of the trajectory and out of frame information at the end.

We inspect the results of the dart across the time intervals of $0.04\,\text{s}$-$0.03\,\text{s}$, $0.03\,\text{s}$-$0.02\,\text{s}$,  $0.02\,\text{s}$-$0.01\,\text{s}$ and $0.01,\text{s}$-$0\,\text{s}$.
Across each interval for both training and testing we are within the discretization interval, showing good prediction performance.
We also observe a similar pattern to the ball dataset in $\theta$: we find our estimate increases in accuracy for smaller time to collision values up until the last time to collision range.
A reason for the increase in error for the last time to collision range is the object may move out of the field of view of the camera and have less spatiotemporal features for predictions. Nonetheless, we still identify this inherent speed accuracy trade off in our dart model.

\subsection{Architecture Design}
\begin{figure}[t]
  \centering
  \includegraphics[width=\linewidth]{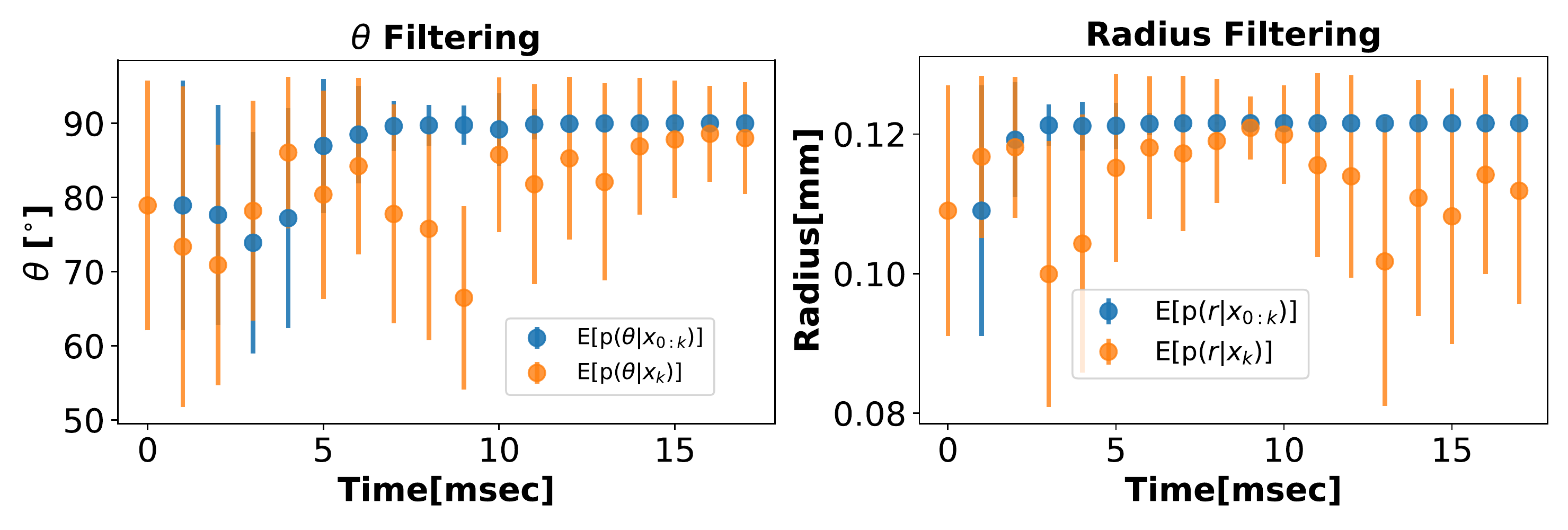}
  \caption{Bayesian filtering of noisy instantaneous probabilities to produce a smoothed impact location estimate of $0.121m$ and $90^{\circ}$. Note the point estimates vary within a discretization interval although with time the filtered probabilities produce a stable estimate.  Error bars represent estimate standard deviation. }
  \label{fig:bayes}
\end{figure}

Many design choices were performed while designing our end to end system architecture.
In this section, we analyze some of the key front end processing decisions such as the importance of augmentation, the exponential filter, and back end decisions such as estimate filtering.

\subsubsection{Exponential Filters}
this subsystem allows us to encode history about past time events and use this for prediction.
To evaluate its importance, we removed the exponential filter from our system architecture and instead relied on present event information by binning the events every $477\,\mu\text{s}$.
After this step, the augmentation procedure was performed and a new \gls{cnn} network with a 2 channel input was trained.
The removal of the exponential filter for the testing set resulted in a $2\times$ error increase in radius, $3 \times$ error increase in $\theta$ and $1.68 \times$ increase in median time to collision error compared to testing results with the exponential filter, as shown by Fig.~\ref{fig:ablation}.
These results highlight across all metrics the exponential filter enhances prediction results.
One reason for the drop in performance is the network loses some temporal information important for prediction.

\subsubsection{Augmentations}
\begin{figure}[t]
  \centering
  \includegraphics[width=\linewidth]{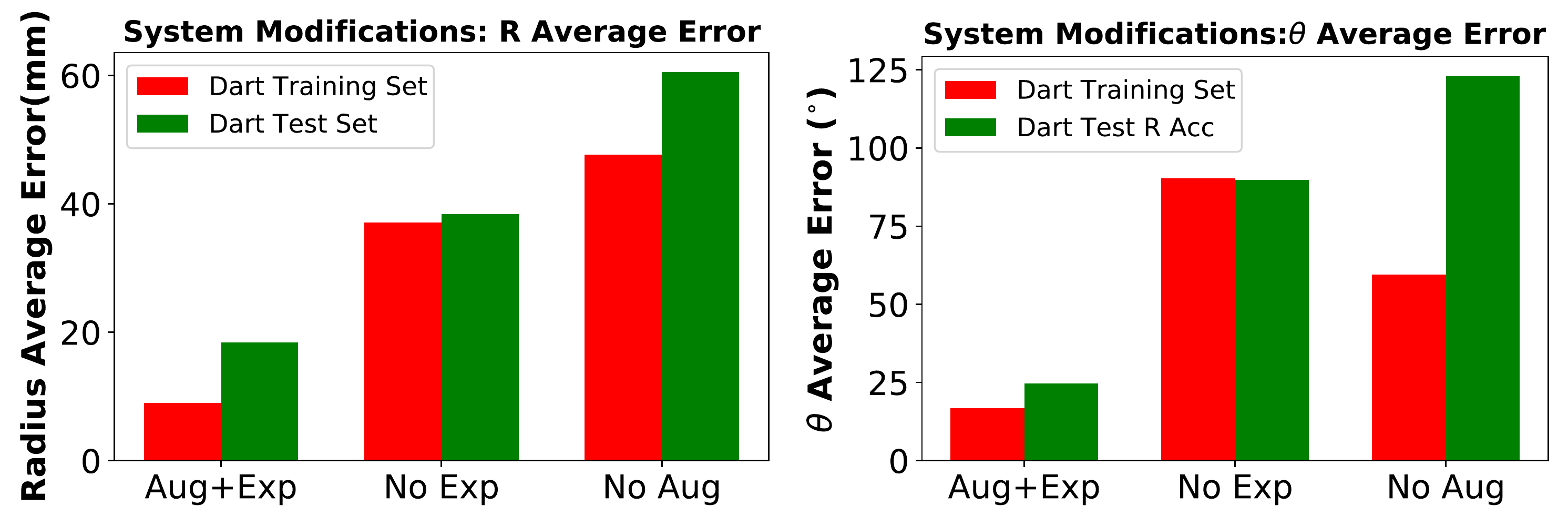}
  \caption{The role of the architecture components (Aug: Augmentation, Exp: Exponential Filter) on the dart dataset, compared to performing both (Aug+Exp) removing the exponential filter (No Exp) results in a increase of $2\times$ radius error and $3\times$ $\theta$ error. Also, compared to performing both (Aug+Exp), removing the augmentation stage results in a increase of $2.74\times$ radius error and $5.1\times$ $\theta$ error. }
  \label{fig:ablation}
\end{figure}

they allow our model to understand further generated trajectories and to prevent overfitting to our dataset.
The next set of experiments we studied was the removal of augmentations from the dart dataset.  We trained on our limited set of $36$ dart trajectories. As a result, our network testing error compared to using augmentations, increased by $2.74 \times$ for radius, increased by $5.1 \times$ for $\theta$ and increased by $30 \times$ for time to collision (Fig. \ref{fig:ablation}). This drop in performance highlights the importance of augmentations for accurate estimation.

\subsubsection{Bayesian Filter}
one study we performed to improve the accuracy of our network's impact location predictions was performing a Bayesian filter on the network output (Fig. \ref{fig:bayes}).
This procedure consists of filtering the softmax probabilities of radius and theta, using the previous time history as the prior for the next posterior update.
We highlight the global dart prediction results for integrating in time to collision ranges of $T_{start}$-0.03, $T_{start}$-0.02, $T_{start}$-0.01, where $T_{start}$ is the beginning time of the trajectory.  One pattern that emerges from these results is we see a $30 \%$ reduction in $\theta$ error and $51\%$ reduction radius between the shortest and longest filtered estimates.
Again, this highlights a speed accuracy trade-off in our model between making a quicker decision versus getting a more accurate impact location prediction.

\subsection{Baselines}
We compared our approach against various baseline methods. The first baseline method was using a RealSense depth camera in order to detect the incoming toy dart and predict its motion.
The RealSense camera was configured for capturing   $480 \times 640$ images at $90\,\text{Hz}$.  We shot a dart directly at the center of the RealSense and recorded the depth and RGB frames. Our result showed the depth camera was unable to detect the dart due to motion blur and its small size. The RGB imager was able to receive only 3 very blurred camera frames of the dart during this time.
Since the RGB images are not adequate for optical flow based tracking, we next looked into using the aggregated DVS images to compute optical flow.

Our next set of baselines considered using optical flow based methods on $3\,\text{ms}$ binned \gls{dvs} events.
We first tried to perform feature based tracking methods on the ball in the image.
This was not successfull because very few points were tracked on the object and some features spuriously tracked noise in the image. Hence, we tried more robust circle tracking methods of the ball via Hough transformations.
The challenge in using the Hough transform was that the large scale values produced the strongest response.
As a result, the true circle diameter was underestimated resulting in very inaccurate motion predictions.
One final approach was creating a convex hull out of the positive and negative events in the circle region. We then calculated the effective diameter given the hull area. This approach solved the issue of biased diameter estimates, yet resulted in large variances.
All in all, applying both model based and feature tracking methods to event based camera were not robust. These results were performed for the simpler case of tracking a ball and would be even worse for faster moving objects, such as the dart.

%%%%%%%%%%%%%%%%%%%%%%%%%%%%%%%%%%%%%%%%%%%%%%%%%%%%%%%%%%%%%%%%%%%%%%%%%%%%%%%%

%%%%%%%%%%%%%%%%%%%%%%%%%%%%%%%%%%%%%%%%%%%%%%%%%%%%%%%%%%%%%%%%%%%%%%%%%%%%%%%%

\section{Conclusion and Future Work}

This paper considered the problem of estimating the collision time and impact location of a fast approaching object, such as a fast moving dart. These pose considerable challenges to conventional cameras and associated motion estimation techniques.

To circumvent these challenges, we focused on Dynamic Vision Sensors and presented a novel motion estimation architecture which encodes the event volume using a bank of  exponential filters.
To train the system, we developed novel  augmentation techniques for event data which use three dimensional motion data.
We showed that the resulting system can detect a dart moving at \nerfmaxspeed{} with a \dartdegerror{} error in $\theta$, \dartraderror{} average radius prediction error and \collisionerror{} median time to collision prediction error. These results constitute the first DVS based system which can reason about objects moving at such high speeds.

There are a number of directions for future work. In the short term, we would like to extend the set of objects in our dataset and consider multiple moving objects. Ultimately we are interested in scenarios where both the camera and approaching object are moving at fast rates and where motion understanding will require processing the input event stream with additional sensors such as inertial measurements.

\addtolength{\textheight}{-12cm}   % This command serves to balance the column lengths

%%%%%%%%%%%%%%%%%%%%%%%%%%%%%%%%%%%%%%%%%%%%%%%%%%%%%%%%%%%%%%%%%%%%%%%%%%%%%%%%

\bibliographystyle{IEEEtran}
\bibliography{IEEEabrv, icra}

\begin{thebibliography}{10}
\providecommand{\url}[1]{#1}
\csname url@samestyle\endcsname
\providecommand{\newblock}{\relax}
\providecommand{\bibinfo}[2]{#2}
\providecommand{\BIBentrySTDinterwordspacing}{\spaceskip=0pt\relax}
\providecommand{\BIBentryALTinterwordstretchfactor}{4}
\providecommand{\BIBentryALTinterwordspacing}{\spaceskip=\fontdimen2\font plus
\BIBentryALTinterwordstretchfactor\fontdimen3\font minus
  \fontdimen4\font\relax}
\providecommand{\BIBforeignlanguage}[2]{{%
\expandafter\ifx\csname l@#1\endcsname\relax
\typeout{** WARNING: IEEEtran.bst: No hyphenation pattern has been}%
\typeout{** loaded for the language `#1'. Using the pattern for}%
\typeout{** the default language instead.}%
\else
\language=\csname l@#1\endcsname
\fi
#2}}
\providecommand{\BIBdecl}{\relax}
\BIBdecl

\bibitem{Pretto_2020}
A.~{Pretto}, S.~{Aravecchia}, W.~{Burgard}, N.~{Chebrolu}, C.~{Dornhege},
  T.~{Falck}, F.~V. {Fleckenstein}, A.~{Fontenla}, M.~{Imperoli}, R.~{Khanna},
  F.~{Liebisch}, P.~{Lottes}, A.~{Milioto}, D.~{Nardi}, S.~{Nardi},
  J.~{Pfeifer}, M.~{Popovic}, C.~{Potena}, C.~{Pradalier},
  E.~{Rothacker-Feder}, I.~{Sa}, A.~{Schaefer}, R.~{Siegwart}, C.~{Stachniss},
  A.~{Walter}, W.~{Winterhalter}, X.~{Wu}, and J.~{Nieto}, ``{Building an
  Aerial-Ground Robotics System for Precision Farming: An Adaptable
  Solution},'' \emph{IEEE Robotics Automation Magazine}, 2020.

\bibitem{chen2019sloam}
S.~W. {Chen}, G.~V. {Nardari}, E.~S. {Lee}, C.~{Qu}, X.~{Liu}, R.~A.~F.
  {Romero}, and V.~{Kumar}, ``{SLOAM: Semantic Lidar Odometry and Mapping for
  Forest Inventory},'' \emph{IEEE Robotics and Automation Letters}, vol.~5,
  no.~2, pp. 612--619, 2020.

\bibitem{tabib2020autonomous}
W.~{Tabib}, K.~{Goel}, J.~{Yao}, C.~{Boirum}, and N.~{Michael}, ``{Autonomous
  Cave Surveying with an Aerial Robot},'' \emph{arXiv:2003.13883}, 2020.

\bibitem{wang2017characterization}
Z.~{Wang}, Y.~{Liu}, Q.~{Liao}, H.~{Ye}, M.~{Liu}, and L.~{Wang},
  ``{Characterization of a RS-LiDAR for 3D Perception},'' in \emph{2018 IEEE
  8th Annual International Conference on CYBER Technology in Automation,
  Control, and Intelligent Systems (CYBER)}, 2018, pp. 564--569.

\bibitem{Zaffar_2018}
M.~{Zaffar}, S.~{Ehsan}, R.~{Stolkin}, and K.~M. {Maier}, ``{Sensors, SLAM and
  Long-term Autonomy: A Review},'' in \emph{2018 NASA/ESA Conference on
  Adaptive Hardware and Systems (AHS)}, 2018, pp. 285--290.

\bibitem{OpticalFlowHistoric}
J.~L. {Barron}, D.~J. {Fleet}, S.~S. {Beauchemin}, and T.~A. {Burkitt},
  ``{Performance of optical flow techniques},'' in \emph{Proceedings 1992 IEEE
  Computer Society Conference on Computer Vision and Pattern Recognition},
  1992, pp. 236--242.

\bibitem{drosophilia}
G.~{Card} and M.~H. {Dickinson}, ``{Visually Mediated Motor Planning in the
  Escape Response of Drosophila},'' \emph{Current Biology}, vol.~18, no.~17,
  pp. 1300 -- 1307, 2008.

\bibitem{Gallego_2020}
G.~{Gallego}, T.~{Delbruck}, G.~M. {Orchard}, C.~{Bartolozzi}, B.~{Taba},
  A.~{Censi}, S.~{Leutenegger}, A.~{Davison}, J.~{Conradt}, K.~{Daniilidis},
  and D.~{Scaramuzza}, ``{Event-based Vision: A Survey},'' \emph{IEEE
  Transactions on Pattern Analysis and Machine Intelligence}, pp. 1--1, 2020.

\bibitem{obstacleAvoidanceRadar}
M.~P. {Owen}, S.~M. {Duffy}, and M.~W.~M. {Edwards}, ``{Unmanned aircraft sense
  and avoid radar: Surrogate flight testing performance evaluation},'' in
  \emph{2014 IEEE Radar Conference}, 2014, pp. 0548--0551.

\bibitem{ousterLidar}
\BIBentryALTinterwordspacing
{Ouster, Inc.}, ``{Mid-Range High-Resolution Imaging Lidar - Specification
  Sheet},'' 2020, accessed October 2020. [Online]. Available:
  \url{http://data.ouster.io/downloads/OS1-lidar-sensor-datasheet.pdf}
\BIBentrySTDinterwordspacing

\bibitem{SGMFpga}
O.~{Rahnama}, T.~{Cavalleri}, S.~{Golodetz}, S.~{Walker}, and P.~{Torr},
  ``{R3SGM: Real-Time Raster-Respecting Semi-Global Matching for
  Power-Constrained Systems},'' in \emph{2018 International Conference on
  Field-Programmable Technology (FPT)}, 2018, pp. 102--109.

\bibitem{stereoVisionQuad}
K.~{McGuire}, G.~{de Croon}, C.~{De Wagter}, K.~{Tuyls}, and H.~{Kappen},
  ``{Efficient Optical Flow and Stereo Vision for Velocity Estimation and
  Obstacle Avoidance on an Autonomous Pocket Drone},'' \emph{IEEE Robotics and
  Automation Letters}, vol.~2, no.~2, pp. 1070--1076, 2017.

\bibitem{monocularVision}
J.~{Michels}, A.~{Saxena}, and A.~Y. {Ng}, ``{High Speed Obstacle Avoidance
  Using Monocular Vision and Reinforcement Learning},'' in \emph{Proceedings of
  the 22nd International Conference on Machine Learning}, ser. ICML '05.\hskip
  1em plus 0.5em minus 0.4em\relax New York, NY, USA: Association for Computing
  Machinery, 2005, p. 593–600.

\bibitem{davideMicroDrone}
D.~{Palossi}, A.~{Loquercio}, F.~{Conti}, E.~{Flamand}, D.~{Scaramuzza}, and
  L.~{Benini}, ``{A 64-mW DNN-Based Visual Navigation Engine for Autonomous
  Nano-Drones},'' \emph{IEEE Internet of Things Journal}, vol.~6, no.~5, pp.
  8357--8371, 2019.

\bibitem{ballDetection}
A.~{Glover} and C.~{Bartolozzi}, ``{Event-driven ball detection and gaze
  fixation in clutter},'' in \emph{2016 IEEE/RSJ International Conference on
  Intelligent Robots and Systems (IROS)}, 2016, pp. 2203--2208.

\bibitem{evdodgeNet}
N.~J. {Sanket}, C.~M. {Parameshwara}, C.~D. {Singh}, A.~V. {Kuruttukulam},
  C.~{Fermüller}, D.~{Scaramuzza}, and Y.~{Aloimonos}, ``{EVDodgeNet: Deep
  Dynamic Obstacle Dodging with Event Cameras},'' in \emph{2020 IEEE
  International Conference on Robotics and Automation (ICRA)}, 2020, pp.
  10\,651--10\,657.

\bibitem{zhu2019eventgan}
A.~Z. Zhu, Z.~Wang, K.~Khant, and K.~Daniilidis, ``{EventGAN: Leveraging Large
  Scale Image Datasets for Event Cameras},'' \emph{arXiv preprint
  arXiv:1912.01584}, 2019.

\bibitem{davideObstacleAv}
D.~{Falanga}, K.~{Kleber}, and D.~{Scaramuzza}, ``{Dynamic Obstacle Avoidance
  for Quadrotors with Event Cameras},'' \emph{Science Robotics}, vol.~5,
  no.~40, 2020.

\bibitem{cameraCalib}
\BIBentryALTinterwordspacing
Robotics and P.~Group, ``{DVS Camera Calibration},'' accessed October 2020.
  [Online]. Available:
  \url{https://github.com/uzh-rpg/rpg_dvs_ros/tree/master/dvs_calibration}
\BIBentrySTDinterwordspacing

\end{thebibliography}

\end{document}